\documentclass{article}

\usepackage{microtype}
\usepackage{graphicx}
\usepackage{subcaption}
\usepackage{booktabs} 
\usepackage{amssymb, commath, bm, amsbsy}

\usepackage{multirow}

\usepackage{hyperref}

\usepackage[accepted]{icml2019}

\icmltitlerunning{Discriminative Few-Shot Learning Based on Directional Statistics}

\begin{document}

\twocolumn[
    \icmltitle{Discriminative Few-Shot Learning Based on Directional Statistics}
    
    
    
    \icmlsetsymbol{equal}{*}
    
    \begin{icmlauthorlist}
    \icmlauthor{Junyoung Park}{tbrain}
    \icmlauthor{Subin Yi}{tbrain}
    \icmlauthor{Yongseok Choi}{tbrain}
    \icmlauthor{Dong-Yeon Cho}{tbrain}
    \icmlauthor{Jiwon Kim}{tbrain}
    \end{icmlauthorlist}
    
    \icmlaffiliation{tbrain}{SK T-Brain, Seoul, South Korea}
    
    \icmlcorrespondingauthor{Junyoung Park}{jypark@sktbrain.com}
    
    
    \vskip 0.3in
]



\printAffiliationsAndNotice{}  

\begin{abstract}
Metric-based few-shot learning methods try to overcome the difficulty due to the lack of training examples by learning embedding to make comparison easy. We propose a novel algorithm to generate class representatives for few-shot classification tasks. As a probabilistic model for learned features of inputs, we consider a mixture of von Mises-Fisher distributions which is known to be more expressive than Gaussian in a high dimensional space. Then, from a discriminative classifier perspective, we get a better class representative considering inter-class correlation which has not been addressed by conventional few-shot learning algorithms. We apply our method to \emph{mini}ImageNet and \emph{tiered}ImageNet datasets, and show that the proposed approach outperforms other comparable methods in few-shot classification tasks.
\end{abstract}

\section{Introduction}
Usually, huge volumes of data are required to train deep neural networks for target applications such as image classification, speech recognition and machine translation. Moreover, in case that new data sets are given, conventional deep learning methods start training networks from scratch. This may take days to weeks with many high performance devices like GPUs. Using continually accumulated information by study or experience, on the other hand, humans are able to learn novel tasks much more efficiently with a few examples or even just one example \cite{Carey78}.

One of the attempts to bridge the gap between the human learning and deep learning is \emph{meta-learning} or \emph{learning-to-learn} \cite{Thrun98}, where neural networks are trained to quickly adapt to new environments or solve unseen tasks during the training phase with a limited number of examples. Few-shot regression and classification have been considered as typical tasks of meta-learning in the supervised learning domain and addressed in many different ways such as recurrent model-based methods \cite{Munkhdalai17a,Mishra18}, optimization-based methods \cite{Ravi17,Finn17}, metric-based methods \cite{Vinyals16,Snell17}, and the variants or combinations of these ones \cite{Lee18}. Most above approaches adopted episodic training framework where a collection of tasks (e.g., {\it N} different classes, {\it K} examples sampled from each class) was considered as a training data set in each learning step without pre-trained large networks utilized in recently proposed methods \cite{Oreshkin18,Gidaris18, Qiao18}.

Similar to nearest neighbor algorithms, metric-based approaches seem to be the simplest and most efficient way to solve few-shot learning. In \emph{Prototypical Networks} \cite{Snell17}, for example, a good representation space is learned so that each class has one representative called a prototype, and examples of that class are clustered around the corresponding prototype in the learned space. Then, new examples are classified by choosing the nearest prototype. In the representation space, a prototype for a class is defined as a mean of examples which belong to the corresponding class, and this choice is justified when distances are computed from a Bregman divergence \cite{Banerjee05-2}.

Our method is based on the same inductive bias, that is, such prototypes are assumed to be in the embedding space. We consider a \emph{von Mises-Fisher} (vMF) mixture model on the learned feature space, and choose prototypes from a discriminative model point of view. Our approach resorts to popular belief that discriminiative models usually solve classification problems better than generative models. However, it is hard to obtain a closed-form solution for discriminative parameters in our setting. Thus, we derived an approximated solution with an additional neural network.

The reason for choosing the vMF distribution other than the Gaussian distribution is that, for the vMF, we have only one controllable parameter $\kappa$ which acts as a variance. For a Gaussian distribution, the number of parameters in its covariance matrix is proportional to the square of the number of variables. In addition, the vMF distributions are known to be more expressive than the Gaussian in a high dimensional space. We can also observe that the naturally induced classifiers based on the vMF mixture model have a form of a softmax over scaled cosine similarity, which was successfully applied in recent few-shot learning literature \cite{Gidaris18, Qiao18, Qi18}.

In this paper, we propose a new metric-based few-shot learning algorithm: (i) based on directional statistics realized by a mixture of the vMF distributions, (ii) from a discrimination perspective considering inter-class correlation which has not been addressed by conventional algorithms. In summary, our key contribution is to propose a novel prototype generator based on a theoretical basis with vMF mixture model viewpoint. We show the effectiveness of the proposed algorithm on two widely used few-shot classification benchmarks, \emph{mini}ImageNet and \emph{tiered}ImageNet.

The remainder of the paper is organized as follows. In Section 2, we summarize related work. With notation of Section 3, we describe the proposed algorithm in Section 4. In Section 5, experimental results are givien. Finally, we conclude and offer some directions for future research in Section 6.

\section{Related Work}
Metric-based few-shot learning aims to learn a feature representation in such a way that samples from the same category could be clustered together in the learned representation space. Matching Networks \cite{Vinyals16} use attention LSTMs to create the contextual embedding considering both a few labeled examples (i.e., support set) and new instances. Prototypical Networks \cite{Snell17} assume that each class has one prototype and a classifier can assign classes to given samples by choosing the nearest prototype. Our method is based on the similar idea, but the prototypes are created differently on a theoretical basis for the use of the von Mises-Fisher (vMF) distribution from the discriminative approach point of view. In Relation Network \cite{Sung18}, there are two neural networks; one for general embedding and the other for metric learning. Our method uses an additional neural network to learn a metric, but it is used to measure the distance between support examples only while distances between support and query ones are measured in Relation Network.

Pre-trained neural networks can be utilized to supply not only the appropriate embedding but also a good configuration for final classifiers. In \cite{Qiao18}, activations from the pre-trained networks are used to train another network whose outputs are predicted parameters for the last layer. In \cite{Gidaris18}, a classifier is trained first, then an attention network on the learned parameters is learned to create parameters for novel examples. These ideas have similarity to our method, but we use a neural network to indirectly create parameters through an induced formula. Moreover, our method does not resort to the pre-training.

Some recent few-shot learning methods adopt scaled cosine similarity for the final softmax layer. In \cite{Gidaris18}, the cosine similarity is used to overcome the difference in weight magnitudes of seen and novel categories.
In \cite{Qi18}, the cosine similarity based recognition model is trained
with scaling factors because the softmax function whose output is in $[-1, 1]$ cannot produce a ground-truth one-hot distribution. The importance of scaling factor is also explained in \cite{Oreshkin18} by observing how the gradients change when the scale factor becomes close to extreme cases. We show that the form of scaled cosine similarity can be naturally induced from a vMF mixture model in few-shot learning cases (Section \ref{section:methodology}).

Finally, it is known that directional statistics are better at modeling directional data \cite{Mardia75}. It is observed that the vMF distribution is more appropriate prior than the Gaussian distribution for learning representations of hyperspherical data \cite{Davidson18}. The vMF distribution also has been used for various image classification tasks \cite{Hasnat17, Wang18, Zhe18}, clustering \cite{Banerjee05, Gopal14}, and machine translation \cite{Kumar18}.












\section{Preliminaries}
We consider the standard episodic few-shot learning setting (i.e., $K$-way $N$-shot). Each episode consists of a support set $S=\{(\bm{x}_i^s, y_i^s)\mid i=1, \ldots, NK\}$ and a query set $Q=\{(\bm{x}_j^q, y_j^q)\mid j=1,\ldots,m\}$, where $\bm{x}_i^s, \bm{x}_j^q\in \mathbb{R}^D$ are $D$-dimensional examples and $y_i^s, y_j^q\in \{1, \ldots, K\}$ denote the corresponding classes. Let $\chi_k$ be the set of examples of class $k$ from $S$, i.e., $\chi_k=\{\bm{x}_i^s\mid y_i^s=k\}$.



\textbf{Von Mises-Fisher Distribution}
On the $(p-1)$-dimensional hypersphere $S^{p-1}=\{\bm{x}\in \mathbb{R}^p\mid \Vert \bm{x} \Vert =1\}$, the \emph{von Mises-Fisher distribution} (vMF) is defined as
\begin{equation*}
    f_p(\bm{x}; \bm{\mu}, \kappa) = C_p(\kappa)\exp(\kappa \bm{\mu}^T \bm{x}),
\end{equation*}
where $\bm{\mu}\in \mathbb{R}^p, \Vert \bm{\mu} \Vert=1$ is called the \emph{mean direction}, $\kappa\ge 0$ is the \emph{concentration parameter}, and $C_p(\kappa)$ is the normalizing constant which is equal to
\begin{equation*}
    C_p(\kappa)=\frac{\kappa^{p/2-1}}{(2\pi)^{p/2}I_{p/2-1}(\kappa)}.
\end{equation*}
Here, $I_v$ is the modified Bessel function of the first kind at order $v$ which cannot be written as a closed formula. The vMF distribution shares a property with the multivariate Gaussian distribution in the sense that the maximum entropy density on $S^{p-1}$ subject to the constraint that $E[\bm{x}]$ is fixed is a vMF density \cite{Rao73}.

\section{Methodology} \label{section:methodology}
\begin{figure}[t]
\begin{center}
\centerline{\includegraphics[width=\columnwidth]{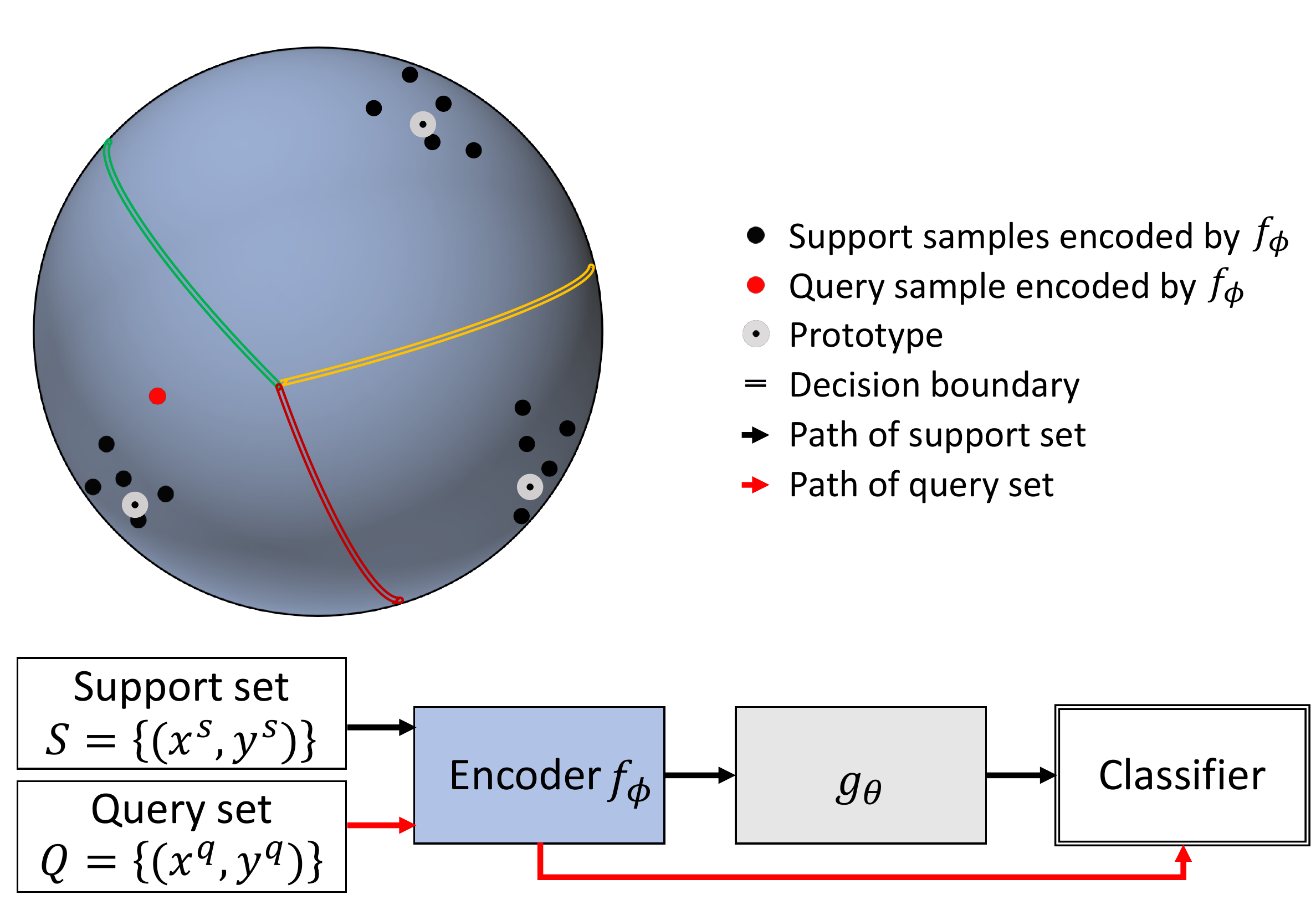}}
\caption{The process for handling an episode in our method. Every element is embedded in the hypersphere through the encoder $f_{\bm{\phi}}$. Then only the embedded support elements are passed as inputs to the function $g_{\theta}$ to generate prototypes for each category. A query element is classified by choosing the nearest prototype. The details are given in Section \ref{section:methodology}.}
\label{figure:our method}
\end{center}
\end{figure}

We assume that there is an encoder $f_{\bm{\phi}}:\mathbb{R}^D\to \mathbb{R}^M$ that maps samples to a representation space. With abuse of notation, we write $\bm{x}$ instead of $L_2$-normalized $f_{\bm{\phi}}(\bm{x})$ in this section for simplicity. The key concept of our method and process for handling an episode are illustrated in Figure \ref{figure:our method}.

\subsection{vMF Mixture Model and Generative Parameters}
First of all, we try to find generative parameters for few-shot classification tasks using a mixture of vMFs. Suppose each class $k$ has its own vMF distribution $f_M(\bm{x}; \bm{\mu}_k, \kappa_k)$ that generates examples of class $k$. Let $z$ be the latent variable for $\bm{x}$ so that $z=k$ if $\bm{x}$ is from class $k$. With a prior $p(z)$, we can write
\begin{align*}
    p(\bm{x}) &= \sum_k p(z=k)p(\bm{x}\mid z=k) \nonumber\\
            &= \sum_k p(z=k)f_M(\bm{x}; \bm{\mu}_k, \kappa_k)
\end{align*}
which represents a mixture of vMFs.

In the episodic few-shot learning, we know all the labels of the support set elements during the training phase. Thus, for a support element $(\bm{x}_i, y_i)=(\bm{x}_i^s, y_i^s)$,
\begin{equation*}
    p(\bm{x}_i\mid z_i=y_i) = f_M(\bm{x}_i;\bm{\mu}_{y_i}, \kappa_{y_i})
\end{equation*}
by the assumption.

Since we consider the same number of examples from each class in few-shot learning settings, it is reasonable to assume that we have uniform prior $p(z)$. For simplicity, we also assume all the $\kappa_k$ coincides, i.e., $\kappa_k = \kappa$ for all $k$. Then, note that
\begin{align} \label{class distribution}
    p(z_i=y_i\mid \bm{x}_i) &= \frac{p(z_i=y_i)p(\bm{x}_i\mid z_i=y_i)}{\sum_k p(z_i=k)p(\bm{x}_i\mid z_i=k)} \nonumber\\
						&=\frac{p(\bm{x}_i\mid z_i=y_i)}{\sum_k p(\bm{x}_i\mid z_i=k)} \nonumber\\
						&= \frac{C_M(\kappa) \exp(\kappa \bm{\mu}_{y_i}^T\bm{x}_i)}{\sum_k C_M(\kappa)\exp(\kappa \bm{\mu}_k^T\bm{x}_i)} \nonumber\\
						&= \frac{\exp(\kappa \bm{\mu}_{y_i}^T\bm{x}_i)}{\sum_k \exp(\kappa \bm{\mu}_k^T \bm{x}_i)}
\end{align}
which is a softmax over $\kappa$-scaled cosine similarities.

Now, let
\begin{align*}
    p(X\mid Z) = \prod_{i=1}^{NK} p(\bm{x}_i\mid z_i=y_i), \\
    p(Z\mid X) = \prod_{i=1}^{NK} p(z_i=y_i\mid \bm{x}_i).
\end{align*}
We will treat $\kappa$ as a hyperparameter, and try to find the parameters $\{\bm{\mu}_k\}$ that maximizes $p(X\mid Z)$ or $p(Z\mid X)$, which we call \emph{generative parameters} and \emph{discriminative parameters}, respectively. Since we use the vMF distribution, we have the constraints that $\Vert \bm{\mu}_k\Vert =1$ for all $k$. To solve this, we set the following Lagrangians:
\begin{align*}
    \mathcal{L}^{gen}(\{\bm{\mu}_k\}, \{\lambda_k\}) = \log p(X\mid Z) + \sum_k \lambda_k (\bm{\mu}_k^T\bm{\mu}_k - 1), \\
    \mathcal{L}^{dis}(\{\bm{\mu}_k\}, \{\lambda_k\}) = \log p(Z\mid X) + \sum_k \lambda_k (\bm{\mu}_k^T\bm{\mu}_k - 1).
\end{align*}
First, we solve the generative parameters. 
By taking partial derivatives of $\mathcal{L}^{gen}$ with respect to $\{\bm{\mu}_k\}$ and $\{\lambda_k\}$, we have
\begin{align*}
    \frac{\partial \mathcal{L}^{gen}}{\partial \bm{\mu}_k} &= \kappa \sum_{x_i\in \chi_k}\bm{x}_i + 2\lambda_k \bm{\mu}_k,\\
    \frac{\partial \mathcal{L}^{gen}}{\partial \lambda_k} &= \bm{\mu}_k^T \bm{\mu}_k-1.
\end{align*}
By setting the above equations to zero and solving, one can check that the solutions are
\begin{align*}
    \bm{\mu}_k^{gen} = \frac{\sum_{\bm{x}_i\in \chi_k}\bm{x}_i}{\Vert \sum_{\bm{x}_i\in \chi_k}\bm{x}_i\Vert}.
\end{align*}

With the vMF distribution, similar argument was also presented in \cite{Banerjee05}. Note that $\bm{\mu}_k^{gen}$ intuitively represents the mean direction in the hypersphere. This is not exactly same as the mean vector defined in Prototypical networks even if the cosine similarity would be used since $\bm{\mu}_k^{gen}$ is the mean of normalized representations on the hypersphere.


\subsection{Main Algorithm}

Now we try to find the discriminative parameters. As we have done in the previous subsection, we take partial derivatives of $\mathcal{L}^{dis}$ with respect to $\{\bm{\mu}_k\}, \{\lambda_k\}$:
\begin{align*}
    \frac{\partial \mathcal{L}^{dis}}{\partial \bm{\mu}_k} &= \sum_{\bm{x}_i\in \chi_k}\kappa \bm{x}_i - \sum_{i=1}^{NK}\frac{\kappa \exp({\kappa \bm{\mu}_k^T \bm{x}_i})\bm{x}_i}{\sum_{k'}\exp({\kappa \bm{\mu}_{k'}^T\bm{x}_i})} + 2\lambda_k \bm{\mu}_k,\\
    \frac{\partial \mathcal{L}^{dis}}{\partial \lambda_k} &= \bm{\mu}_k^T \bm{\mu}_k-1.
\end{align*}
Again, by setting above equations to zero and solving, we have the following equation for local optima $\{\bm{\mu}_k^{dis}=\hat{\bm{\mu}}_k\}$:
\begin{align} \label{mu}
    \hat{\bm{\mu}}_k &= \pm \text{Norm}\bigg\{\sum_{\bm{x}_i\in \chi_k}\bm{x}_i - \sum_{i=1}^{NK}\frac{\exp(\kappa \hat{\bm{\mu}}_k^T\bm{x}_i)}{\sum_{k'}\exp(\kappa \hat{\bm{\mu}}_{k'}^T\bm{x}_i)}\bm{x}_i\bigg\} \nonumber\\
            &=\pm \text{Norm}\bigg\{\sum_{\bm{x}_i\in \chi_k} \bm{x}_i - \sum_{i=1}^{NK} p(z_i=k\mid \bm{x}_i)\bm{x}_i\bigg\},
\end{align}
where Norm is the normalizing operator to make $\bm{\mu}_k$ a unit vector. The second equality follows from Equation (\ref{class distribution}). Note that now $\hat{\bm{\mu}}_k$ depends not only on examples of class $k$ as in generative parameters, but also on examples of other classes. Strictly speaking, we have two possibilities for the sign in the above equation. Since the generative parameters are means of the corresponding class, we take the plus sign to have positive coefficients on the examples of the corresponding class. Empirically, taking all the minus signs proved to have bad performance.

To use Equation (\ref{mu}), we make a learnable function $g_{\theta}$, whose output is $g_{\theta}(k, i; S)\in \mathbb{R}, 0< g_{\theta}(k, i; S)< 1$ for each $i\in \{1, \ldots, NK\}$ and class $k$, that substitutes $p(z_i=k\mid \bm{x}_i)$. This function is implemented by a neural network and used to approximate $\hat{\bm{\mu}}_k$ by $\tilde{\bm{\mu}}_k$:
\begin{equation}
\label{mu_tild}
    \tilde{\bm{\mu}}_k = \text{Norm}\bigg\{\sum_{\bm{x}_i\in \chi_k}\bm{x}_i - \sum_{i=1}^{NK}g_{\theta}(k, i;S)\bm{x}_i\bigg\}.
\end{equation}
Then, for a query example $\bm{x}$, we can classify $\bm{x}$ with the following class distribution
\begin{equation*}
    p(z=k\mid \bm{x})=\frac{\exp(\kappa \tilde{\bm{\mu}}_{k}^T\bm{x})}{\sum_{k'} \exp(\kappa \tilde{\bm{\mu}}_{k'}^T \bm{x})}.
\end{equation*}
Hence with a query set, we define the loss
\begin{align*}
    L = -\frac{1}{m}\log \bigg\{\prod_{j=1}^m p(z=y_j^q\mid \bm{x}_j^q)\bigg\}
\end{align*}
and jointly train $f_{\bm{\phi}}$ and $g_{\theta}$ by gradient descent. The overall procedure of our method is given in Algorithm \ref{alg:main}.

\begin{algorithm}[tb]
   \caption{Main Algorithm}
   \label{alg:main}
\begin{algorithmic}
   \STATE {\bfseries Input:} An episode made up of a support set $S=\{(\bm{x}_i^s, y_i^s)\mid i=1, \ldots, NK\}$ and a query set $Q=\{(\bm{x}_j^q, y_j^q)\mid j=1,\ldots,m\}$. An overline on a letter means normalization.
   \STATE {\bfseries Output:} Loss $L$.
   \FOR{$k=1$ {\bfseries to} $K$}
   \STATE $\bm{\mu}_k=\sum_{\bm{x}_i\in \chi_k}\overline{f_{\bm{\phi}}(\bm{x}_i^s)} - \sum_{i=1}^{NK}g_{\theta}(k, i;S)\overline{f_{\bm{\phi}}(\bm{x}_i^s)}$
   \STATE $\bm{\mu}_k=\overline{\bm{\mu}_k}$
   \ENDFOR
   \STATE Initialize $L = 0$.
   \FOR{$j=1$ {\bfseries to} $m$}
   \STATE $L = L - \kappa \bm{\mu}_{y_j^q}^T \overline{f_{\bm{\phi}}(\bm{x}_j^q)} + \log\{\sum_{k} \exp(\kappa \bm{\mu}_{k}^T \overline{f_{\bm{\phi}}(\bm{x}_j^q)})\}$
   \ENDFOR
   \STATE $L = L/m$
\end{algorithmic}
\end{algorithm}

\subsection{Description of $g_{\bm{\theta}}$}
Here we describe how to construct the function $g_{\bm{\theta}}$ described in the previous subsection. We will train a neural network $d_{\bm{\theta}}$ such that we expect it to operate like a metric. So $d_{\bm{\theta}}(\bm{x}, \bm{y})$ will output a real number for any $\bm{x}, \bm{y}\in \mathbb{R}^M$ but we do not impose it to have properties of a metric such as symmetry or non-negativity. The detailed architecture of $d_{\bm{\theta}}$ will be given in Section \ref{section:experiments}.

Given such $d_{\bm{\theta}}$, we compute all the pairwise distance $d_{ij}=d_{\bm{\theta}}(\bm{x}_i^s, \bm{x}_j^s)$, and let $e_{ik} = \sum_{\bm{x}_j^s\in \chi_k}d_{ij}$. Finally, we define
\begin{equation} \label{g:def}
    g_{\bm{\theta}}(k, i; S) = \frac{\exp{(-e_{ik})}}{\sum_{k'}\exp{(-e_{ik'})}}.
\end{equation}
The motivation of this definition is that, since $g_{\bm{\theta}}(k,i;S)$ represents $p(z_i=k\mid \bm{x}_i^s)$ as shown in Equation (\ref{mu_tild}), we want $g_{\bm{\theta}}(k,i;S)$ to be larger compared to $g_{\bm{\theta}}(k',i;S)$ if $\bm{x}_i^s\in \chi_k$ and $\bm{x}_i^s\notin \chi_{k'}$. After training the networks, we expect that the features from the images of $f_{\bm{\phi}}$ will be clustered according to classes. Thus, $d_{ij}$ will be relatively small if $\bm{x}_i^s$ and $\bm{x}_j^s$ are from the same class. The similar result will happen to $e_{ik}$, thus we can satisfy our goal by Equation (\ref{g:def}). Note that the property $\sum_k p(z=k\mid \bm{x})=1$ also holds for $g_{\bm{\theta}}$.

\section{Experiments} \label{section:experiments}


The proposed few-shot learning method was compared with strong baselines. To evaluate the performance of different approaches, we considered two widely used benchmark datasets for the few-show classification: \textit{mini}ImageNet \cite{Vinyals16} and \textit{tiered}ImageNet datasets \cite{ren18fewshotssl}.

\begin{table*}
  \caption{Few-shot classification accuracy on the \textit{mini}ImageNet data with 95\% confidence intervals ( \textsuperscript{*}No reported confidence intervals)}
  \label{table:MiniImageNet}
  \vskip 0.15in
  \begin{center}
  \begin{small}
  \begin{sc}
  \begin{tabular}{ lcc}
     \toprule
            &   \multicolumn{2}{c}{5-way} \\
     Model  & 1-shot & 5-shot \\
     \midrule
     MatchingNet \cite{Vinyals16}\textsuperscript{*}
                & $42.4\%$        & $58.0\%$ \\
     MatchingNet FCE \cite{Vinyals16}\textsuperscript{*}
                & $46.6\%$        & $60.0\%$ \\
     Meta-Learner LSTM \cite{Ravi17}
                & $43.44\pm0.77\%$ & $60.60\pm0.71\%$ \\
     ProtoNet \cite{Snell17}
                & $49.42\pm0.78\%$ & $68.20\pm0.66\%$ \\
     RelationNet \cite{Sung18}
                & $50.44\pm0.82\%$ & $65.32\pm0.70\%$ \\
     Ours - Metric1 (M1) & $\mathbf{52.44\pm0.20\%}$ & $\mathbf{68.60\pm0.17\%}$ \\
     Ours - Metric2 (M2) & $52.06\pm0.20\%$ & $67.53\pm0.17\%$ \\
     \bottomrule
  \end{tabular}
  \end{sc}
  \end{small}
  \end{center}
\end{table*}

\begin{table*}
  \caption{Few-shot classification accuracy on the \textit{tiered}ImageNet data with 95\% confidence intervals ( \textsuperscript{*}Our reproduction)} 
  \label{table:TieredImageNet}
  \vskip 0.15in
  \begin{center}
  \begin{small}
  \begin{sc}
  \begin{tabular}{ lcc}
     \toprule
                & \multicolumn{2}{c}{5-way} \\
     Model      & 1-shot & 5-shot \\
     \midrule
     ProtoNet \cite{Snell17}\textsuperscript{*}   
                & $47.93\pm0.67\%$ & $69.20\pm0.57\%$ \\
     Ours - Metric1 (M1) & $52.61\pm0.69\%$ & $69.22\pm0.59\%$ \\
     Ours - Metric2 (M2) & $\mathbf{52.94\pm0.69}\%$ & $\mathbf{69.44\pm0.57\%}$ \\
     \bottomrule
  \end{tabular}
  \end{sc}
  \end{small}
  \end{center}
\end{table*}

\subsection{Network Architecture}

For the encoder network $f_{\bm{\phi}}$, we followed the same architecture used in \cite{Vinyals16, Snell17, Sung18}, which is composed of four convolutional blocks. In the first three blocks, each block is composed of a convolutional layer composed of $64~~3\times3$ filters, followed by a batch normalization layer, a ReLU layer, and a $2\times2$ max pooling layer. The last block only has a convolutional layer and a max pooling layer of the same size to have features laid on the entire hypersphere. In all of our experiments, images of size $3\times84\times84$ are used, so the features encoded by the $f_{\bm{\phi}}$ become 1,600 ($=64\times5\times5$) dimensional vectors.

For the distance metric network $d_{\bm{\theta}}$, we propose two neural networks. The first one, which we call M1, consists of a flatten layer, a substraction layer and a two-layered MLP. For the ordered input pair $(\bm{x}, \bm{y})$, we first flatten each input and the substraction layer outputs $\bm{x} - \bm{y}$. After that, the first fully-connected layer reduces the dimension to 8 and the output is additionally processed by a ReLU layer. Then the output is fed to the the second fully-connected layer which reduces the dimension to 1. This metric network has only $12,808$ parameters, which is only a small increase compared to that of the encoder $f_{\bm{\phi}}$ ($\approx 110,000$).

The second distance metric called M2 was inspired by the Relation Network \cite{Sung18}. The architecture is the same as that of the relation module of Relation Network, which has a concatenation layer in depth, two convolutional blocks as in M1 followed by two fully-connected layers with ReLU of size 8 and 1, respectively. Because of its complexity, this network is heavier than M1 in terms of the number of parameters.

\begin{figure*}[!htb]
    \vskip 0.2in
    \centering
    \captionsetup[subfigure]{labelformat=empty}
    \begin{minipage}{0.1in}
        \rotatebox{90}{\textit{mini}ImageNet}
    \end{minipage}%
    \hspace{0.1in}
	\begin{subfigure}[c]{0.45\linewidth}
        \includegraphics[width=\linewidth]{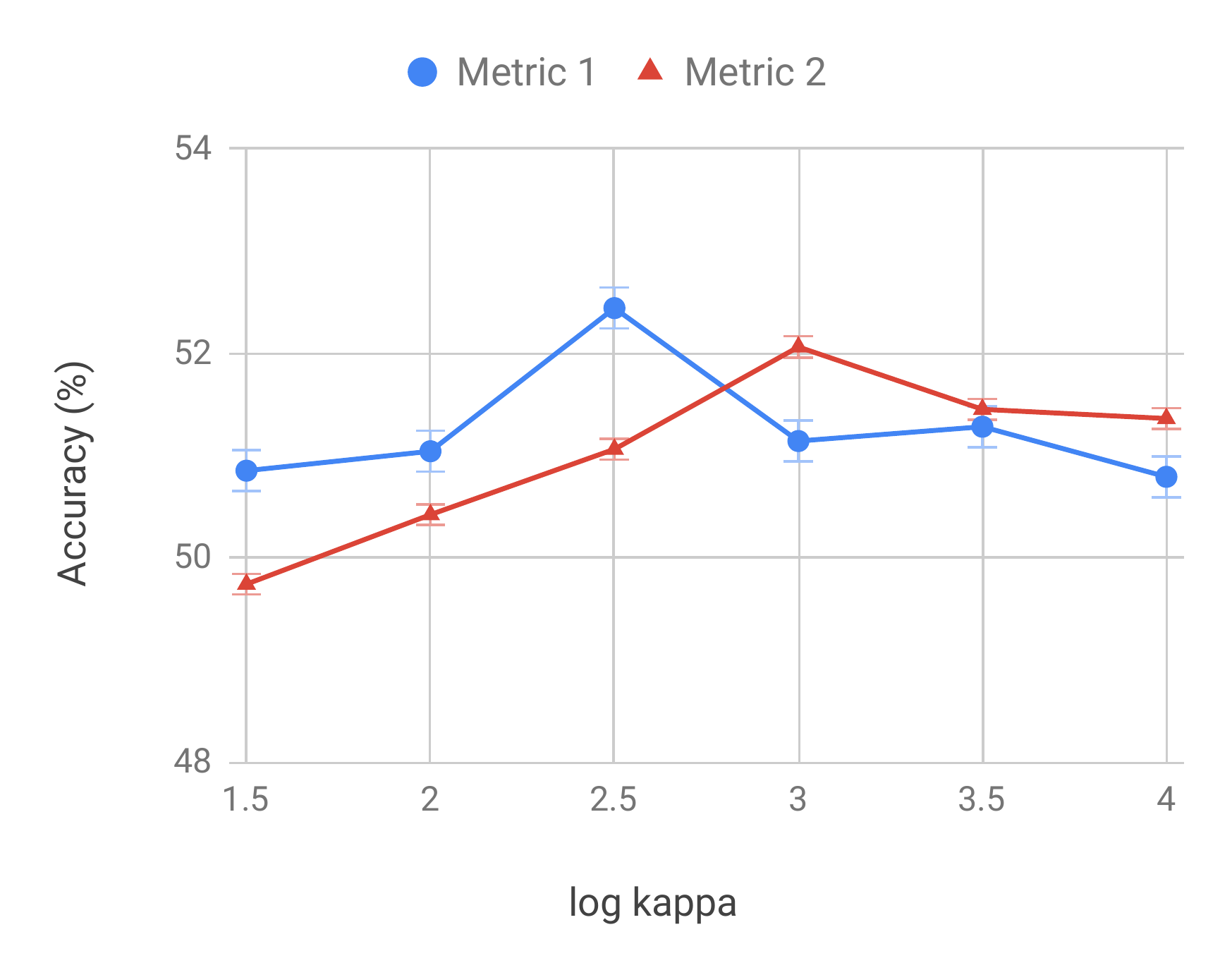}
    \end{subfigure}
    \begin{subfigure}[c]{0.45\linewidth}
        \centering
        \includegraphics[width=\linewidth]{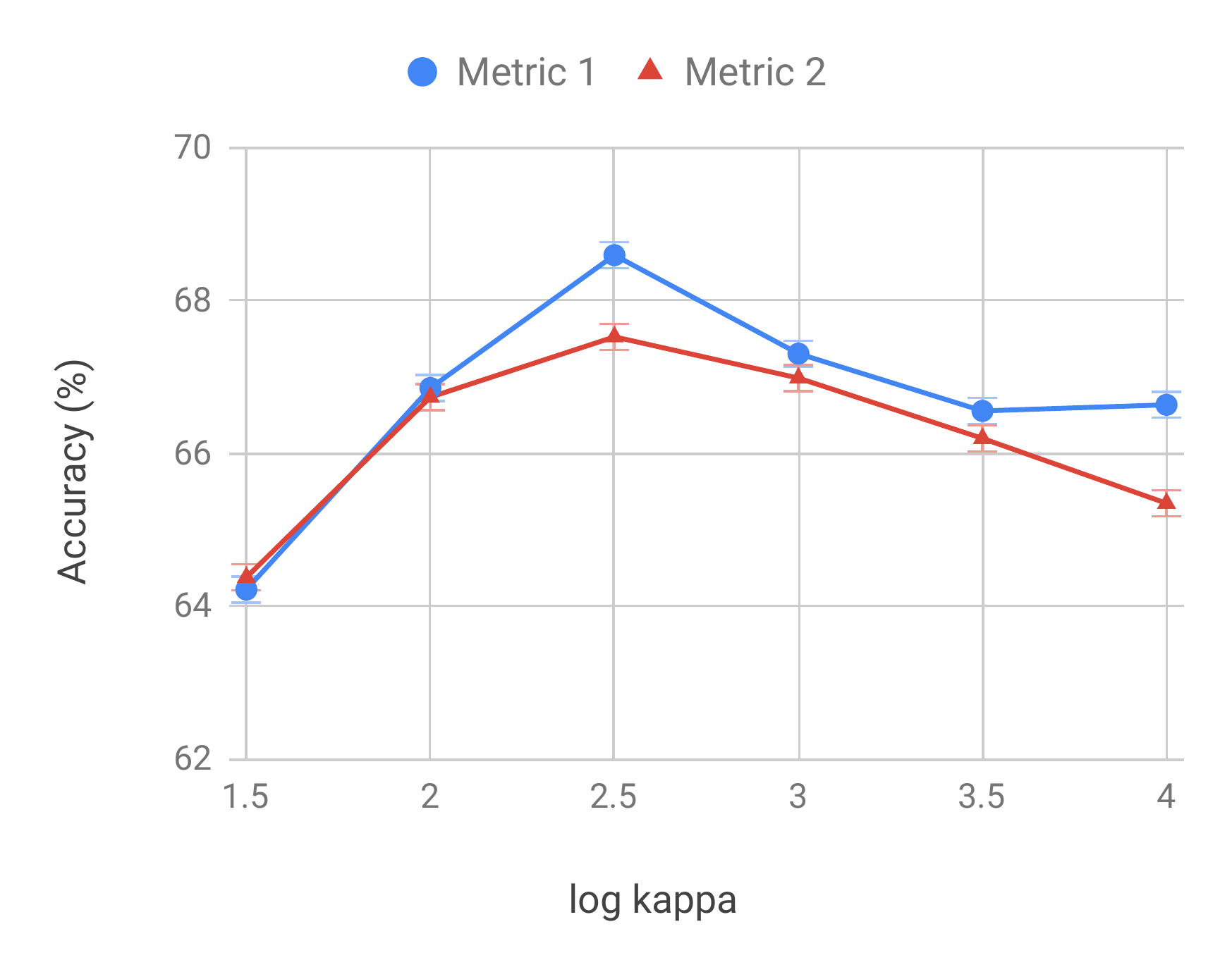}
    \end{subfigure}
        \\
    \begin{minipage}{0.1in}
        \rotatebox{90}{\textit{tiered}ImageNet}
    \end{minipage}%
    \hspace{0.1in}
	\begin{subfigure}[c]{0.45\linewidth}
	    \includegraphics[width=\linewidth]{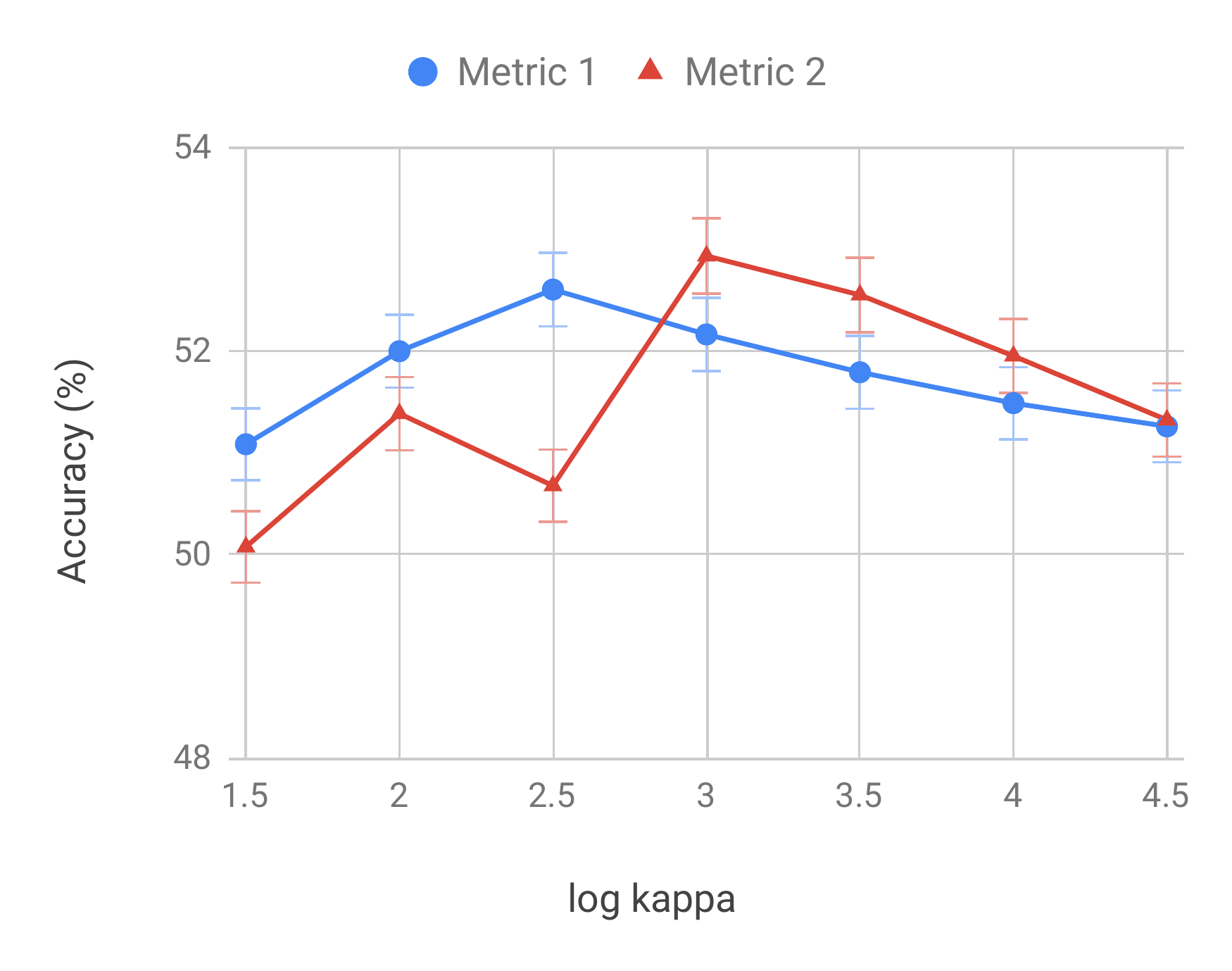}
	    \subcaption{5-way 1-shot}
    \end{subfigure}
    \begin{subfigure}[c]{0.45\linewidth}
        \includegraphics[width=\linewidth]{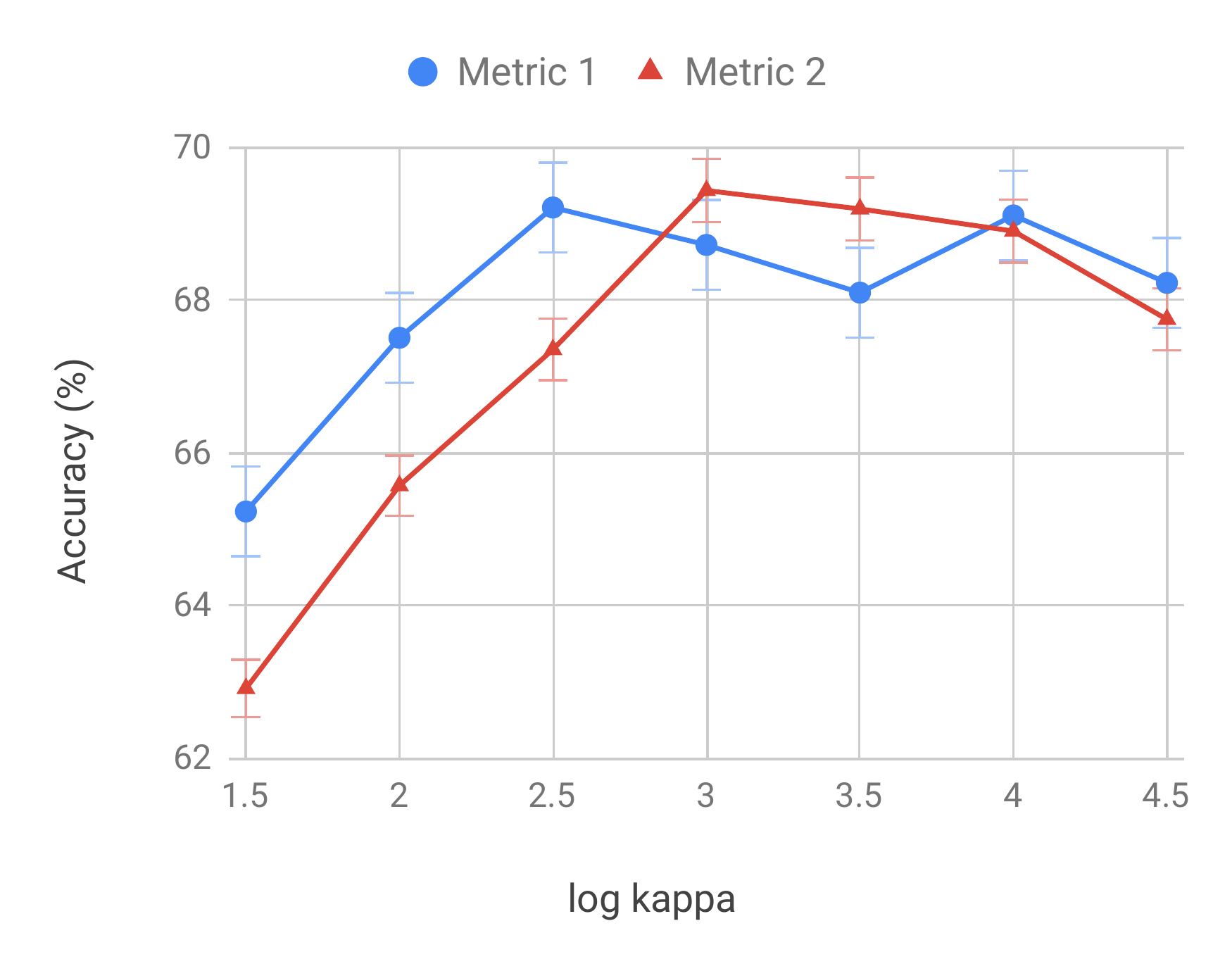}
        \subcaption{5-way 5-shot}
    \end{subfigure}
    \caption{Effects of scaling $\kappa$ on classification accuracy. 
    In case of the \textit{mini}ImageNet experiments, we trained models with metric 1 and metric 2 using 35-way and 30-way episodes for 1-shot classification and 20-way and 15-way episodes for 5-shot classification respectively. In the \textit{tiered}ImageNet experiments, the models with metric 1 and metric 2 were trained using 30-way and 35-way episodes for 1-shot classification and 25-way episodes for 5-shot classification. We matched the train shot and the test shot and sampled 15 queries from each class in every episode.}
    \label{fig:kappa}
    \vskip -0.2in
\end{figure*}

\subsection{\textit{Mini}ImageNet Results}

\textit{Mini}ImageNet dataset, proposed by~\cite{Vinyals16}, is a subset of the ILSVRC-12 ImageNet dataset~\cite{ImageNet:ILSVRC15}. The original ImageNet dataset is a notoriously huge dataset composed of more than a million color images depicting 1,000 object categories, which consumes a large amount of resources to train deep neural networks for their classification. \textit{Mini}ImageNet was proposed to reduce the burden. It contains 60,000 color images of size $84\times84$ from 100 classes, each having 600 examples. We adopted class splits of \cite{Ravi17}. These splits used 64 classes for training, 16 for validation, and 20 for test.


Through $f_{\bm{\phi}}$, we encoded both the support sets and the query sets while only the support sets were used to build the prototypes. All the models were trained via stochastic gradient descent using Adam optimizer with an initial learning rate $10^{-3}$.

We experimented our algorithm on 5-way 1-shot and 5-way 5-shot classification cases. The models with metric 1 (M1) were trained using 35-way episodes for 1-shot classification and 20-way episodes for 5-shot classification with $\log\kappa=2.5$. The models with metric 2 (M2) were also trained using 30-way and 15-way episodes with $\log\kappa=3.0, 2.5$ for 1-shot and 5-shot classification problems, respectively. We used 15 query points for both the train and test data per episode.
The proposed method was compared with recent relevant studies, Matching Network \cite{Vinyals16}, Meta-learner LSTM \cite{Ravi17}, Prototypical Network, \cite{Snell17}, and Relation Network \cite{Sung18}, which had similar base encoder architectures and training strategies as ours, i.e., a four-layer convolutional network with 64 filters without pre-training, for fair comparisons. 

As shown in Table \ref{table:MiniImageNet}, our method had the best performance in both 1-shot and 5-shot experiments with the accuracy 52.44\% and 68.60\%. In case of the 5-shot experiments, the accuracy difference between ours and the Prototypical Network was smaller than the standard deviation, however, ours outperformed the prototypical network with greater accuracy improvement than the standard deviation in case of the 1-shot problems.

\subsection{\textit{Tiered}ImageNet Results}

\textit{Tiered}ImageNet dataset was first proposed by \cite{ren18fewshotssl}. It is also a subset of the ILSVRC-12 ImageNet dataset \cite{ImageNet:ILSVRC15} like the \textit{mini}ImageNet dataset. However, it contains more subsets than the \textit{mini}ImageNet, containing 608 classes whereas the \textit{mini}ImageNet contains 100 classes. In the \textit{Tiered}ImageNet dataset, more than 600 classes are grouped into broader 34 categories which correspond to the nodes located at the higher positions of the ImageNet hierarchy \cite{ImageNetDeng2009}. Each broader category contains 10 to 30 classes. These 34 broader categories are split into 20 train sets, 6 validation sets, and 8 test sets.

Same as in the \textit{mini}ImageNet experiments, the models were trained using Adam optimizer with an initial learning rate $10^{-3}$. 
Models with M1 were trained using 30-way episodes for 1-shot classification and 25-way episodes for 5-shot classification with $\log\kappa=2.5$. Models with M2 were trained using 35-way and 25-way episodes with $\log\kappa=3.0$. We used 15 query points for both the train and test data per episode same as we did in the \textit{mini}ImageNet experiments.

In the \textit{tiered}ImageNet experiments, our method with metric M1 showed the best performance in both 1-shot and 5-shot experiments with the accuracy 52.94\% and 69.44\% as shown in the Table~\ref{table:TieredImageNet}. There were tiny differences between the metric M1 and the metric M2, which were even smaller than the standard deviations. However, our method outperformed Prototypical Networks in both 1-shot and 5-shot problems.

\begin{figure}[tb]
    \vskip 0.2in
    \centering
    \includegraphics[width=0.9\linewidth]{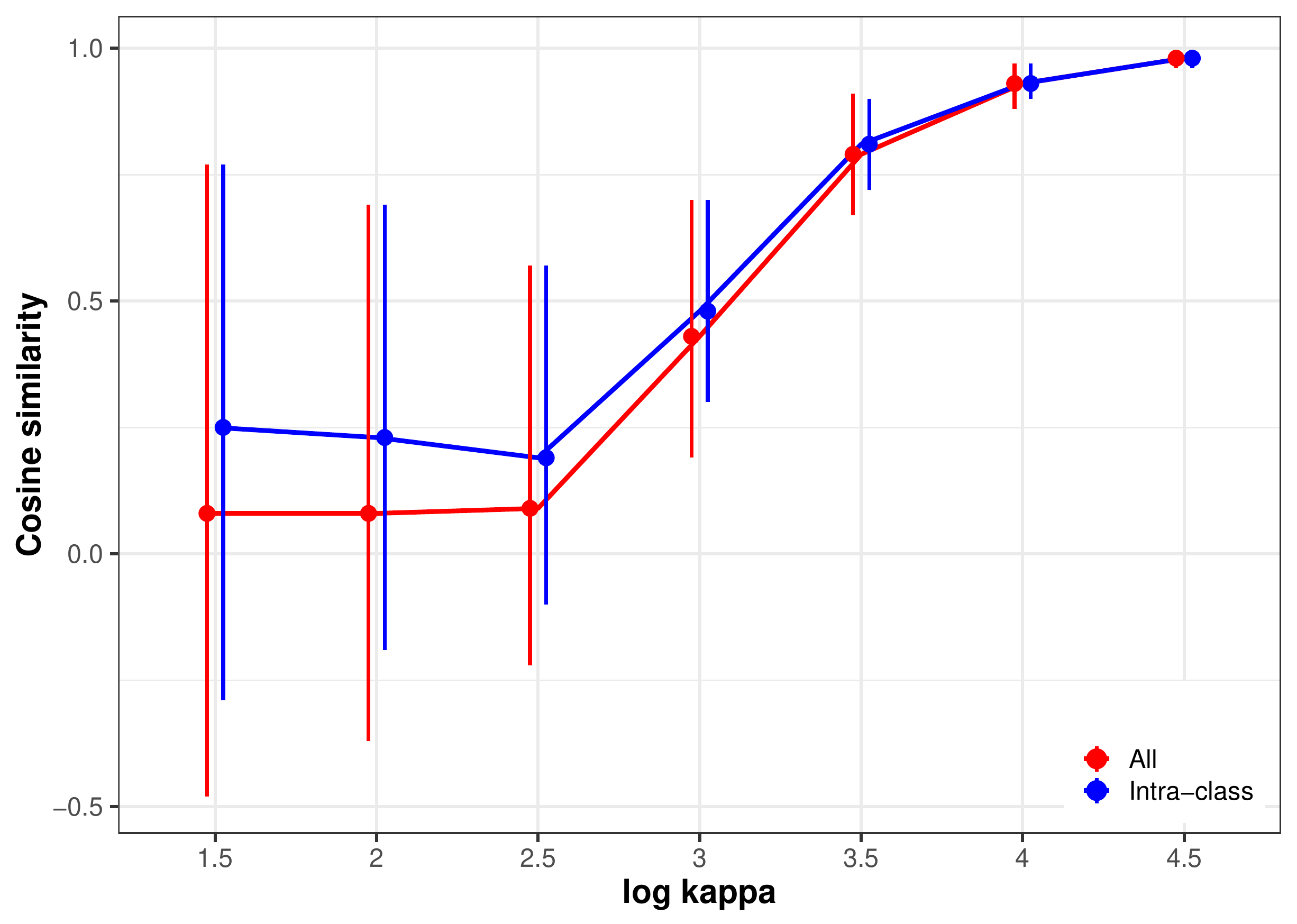} 
    \caption{Changes of pairwise cosine similarities over $\kappa$ in \textit{tiered}ImageNet. A dot in the middle of each bar represents a mean of cosine similarity values for each configuration. The length of each bar represents the range of cosine similarity values (from minimum to maximum values) for $\kappa$.}
    \label{fig:cos_sim_over_kappa}
    \vskip -0.2in
\end{figure}

\subsection{Effects of Scaling the Concentration Parameter $\kappa$}

\subsubsection{Classification Accuracy}

We study how the concentration parameter $\kappa$ of the von Mises-Fisher distribution $f_p(\bm{x};\bm{\mu},\kappa)$ affects the performance of our method. 
The concentration parameter characterizes how strongly the vectors drawn from $f_p(\bm{x};\bm{\mu},\kappa)$ are concentrated around the mean direction $\bm{\mu}$. When $\kappa=0$, $f_p(\bm{x};\bm{\mu},\kappa)$ is equivalent to a uniform density on $S^{p-1}$ and when $\kappa \rightarrow\infty$, $f_p(\bm{x};\bm{\mu},\kappa)$ tends to be a point density. 
Thus, when $\kappa$ is too small, the vMFs of different classes are indistinguishable. On the other hand, when $\kappa$ is too large, only those points that are sufficiently close to mean directions can have decent probability.

Figure \ref{fig:kappa} shows how the classification accuracy changes over various $\kappa$. The graphs have concave forms where the accuracy decreases when $\kappa$ is too small or too big. This implies that there exists an optimal $\kappa$ value which gives the best description of the class distribution and the accuracy reaches the highest point. 

\subsubsection{Configuration of features}

We further explore how $\kappa$ affects the distribution of features which lie on the hypersphere by measuring pairwise cosine similarities of feature vectors. Figure~\ref{fig:cos_sim_over_kappa} shows how cosine similarity values between these feature vectors vary over various $\kappa$ values when applying \textit{tiered}ImageNet test examples to the encoder. 

We can see that as the $\kappa$ increases, the whole features converge to a point, which can be read from a narrower range of the cosine similarities. Although the distribution of features changes drastically according to the value of $\kappa$, the features of a particular class (``intra-class") are more concentrated than those of all classes (``all pairs"), and the mean values of pairwise cosine similarities between examples from the same classes are consistently higher than those of all pairs with a clear distinction. This explains partly why our algorithm shows decent classification accuracy over various $\kappa$ values.

\section{Conclusion}

We have proposed an algorithm for few-shot learning based on the von Mises-Fisher mixture model. To the best of our knowledge, this is the first research that uses vMF distribution on few-shot learning. Furthermore, we propose a novel method that approximates discriminative parameters based on a theoretical basis. Together with an embedding network, our method trains another network whose outputs are prototypes for each episode. Using two standard image datasets, we showed that our method outperformed other baseline approaches that have the similar convolutional encoders as ours in terms of their capacity and training method.

Much deeper encoders such as Wide Residual Networks \cite{ZagoruykoK16} than ours were adopted in recently proposed few-shot learning methods \cite{Rusu18, FEAT:YeHZS2018Learning}. These encoders were usually pre-trained before the few-shot learning phase. Thus, large networks and their pre-training could be considered in our future work.

As shown in our results, the scaling factor (i.e., concentration parameter $\kappa$) has an effect on the classification accuracy. Further improvement can be made by designing an algorithm that automatically chooses the optimal value for this parameter.

\bibliography{reference}
\bibliographystyle{icml2019}





\end{document}